\documentclass[smallextended]{svjour3}

\smartqed  % flush right qed marks, e.g. at end of proof
\usepackage{booktabs}
\usepackage{enumitem}
\usepackage{longtable}
\usepackage{graphicx}
\usepackage{array}
\usepackage{dcolumn}
\usepackage{amsfonts}
\usepackage{amssymb}
\usepackage{xspace}
\usepackage{xfrac}
\usepackage{array}
\usepackage{multirow}
\usepackage{xcolor}
\usepackage{footnote}
\usepackage{url}
\usepackage{makeidx}
\usepackage{epsfig}
\usepackage{amsmath}
\usepackage{multicol}
\usepackage{multirow}
\usepackage{mathtools}
\usepackage[misc]{ifsym}
\usepackage[ruled,linesnumbered]{algorithm2e}
\usepackage{tabularx}
\usepackage{array}
\usepackage{footnote}
\makesavenoteenv{tabular}
\makesavenoteenv{table}
\begin{document}

\title{Automated Imbalanced Classification via Layered Learning}

\titlerunning{Imbalanced Classification Layered Learning}

\author{Vitor~Cerqueira, Luis~Torgo, Paula~Branco, and Colin~Bellinger}

\authorrunning{V. Cerqueira et al.}

\institute{Vitor Cerqueira (\Letter) \at
         Dalhousie University, Halifax, Canada\\
         \email{vitor.cerqueira@dal.ca}
         \and
         Luis~Torgo \at
         Dalhousie University, Halifax, Canada\\
         \and
         Paula~Branco \at
         University of Ottawa, Ottawa, Canada\\
         \and
         Colin Bellinger \at
         National Research Council of Canada, Canada\\
         \and
         Corresponding author: Vitor Cerqueira\\
         }

\date{Received: date / Accepted: date}

\maketitle

\begin{abstract}

In this paper we address imbalanced binary classification (IBC) tasks.
Applying resampling strategies to balance the class distribution of training instances is a common approach to tackle these problems.
Many state-of-the-art methods find instances of interest close to the decision boundary to drive the resampling process. However, under-sampling the majority class may potentially lead to important information loss. Over-sampling also may increase the chance of overfitting by propagating the information contained in instances from the minority class.
The main contribution of our work is a new method called \texttt{ICLL} for tackling IBC tasks which is not based on resampling training observations. Instead, \texttt{ICLL} follows a layered learning paradigm to model the data in two stages.
In the first layer, \texttt{ICLL} learns to distinguish cases close to the decision boundary from cases which are clearly from the majority class, where this dichotomy is defined using a hierarchical clustering analysis. In the subsequent layer, we use instances close to the decision boundary and instances from the minority class to solve the original predictive task.
A second contribution of our work is the automatic definition of the layers which comprise the layered learning strategy using a hierarchical clustering model. This is a relevant discovery as this process is usually performed manually according to domain knowledge.
We carried out extensive experiments using 100 benchmark data sets. The results show that the proposed method leads to a better performance relatively to several state-of-the-art methods for IBC.


\end{abstract}

\section{Introduction}\label{intro}

In supervised learning, we face an imbalanced problem when the distribution of classes is significantly biased towards a particular value and at the same time, the least frequent class is also the most relevant one \cite{branco2016survey}.
An archetype of imbalanced classification is the detection of fraudulent cases, which are rare occurrences among a large proportion of normal activity. The accurate detection of these rare but important instances is fundamental across many domains of application. In effect, learning from imbalanced domains is one of the most active research topics in the machine learning literature.

The skewed class distribution hinders the learning process of algorithms, and several strategies have been devised to overcome this problem.
The large majority of approaches designed to tackle the class imbalance issue are based on resampling methods.
These transform the training data to improve the relevance of the minority class. Examples include SMOTE \cite{chawla2002smote}, ADASYN \cite{he2008adasyn}, among many others \cite{branco2016survey}.
These methods are versatile and easy to couple with any learning system.
However, under-sampling the majority class may potentially lead to important information loss. Over-sampling may increase the chance of overfitting by propagating the information contained in instances from the minority class.

Many of the existing resampling approaches work by finding instances close the decision boundary and using those instances to drive the resampling process. For example, ADASYN \cite{he2008adasyn} is a popular method which synthesizes instances from the minority class whose neighborhood belong to the majority class.
In this paper we propose a novel approach for imbalanced binary classification problems, which is called \texttt{ICLL} (Imbalanced Classification via Layered Learning).
Specifically, we propose a method that, unlike resampling approaches, attempts to capture and improve the modelling of instances close to the decision boundary without synthesizing new instances or removing information from the training set.

First, we consider that an instance is arbitrarily close to the decision boundary according to a hierarchical clustering analysis. After applying the clustering model and cutting the hierarchy using an automatic heuristic \cite{bellinger2019cure}, instances are assigned to one of three groups: the pure majority group, if that instance is clustered with only observations of the majority class; the pure minority group, if the respective cluster contains only observations of the minority class; or the mixed group, if the respective cluster contains observations from both classes. Accordingly, an observation is considered borderline (i.e., close to the decision boundary) if it belongs to the mixed group.

After assigning the training instances to one of these three groups based on the class distribution of the respective cluster, we model the data set using a layered learning approach \cite{stone2000layered}. 
Layered learning represents a hierarchical learning paradigm in which a predictive task is split into multiple, expectedly simpler, sub-tasks.
In this work, we adopt a two-layer strategy following the work by Cerqueira et al. \cite{cerqueira2020early}.
The first layer denotes a binary task designed to distinguish instances of the pure majority group from instances of either the mixed or pure minority group. In other words, we attempt to separate those instances which are clearly from the majority class (i.e. belong to the pure majority group).
The second layer represents the original predictive task, where the objective is to distinguish instances from the majority class from instances of the minority class. The major difference is that only observations from mixed or pure minority groups are considered; instances from the pure majority group are discarded as the system models them in the first layer. Inference is performed according to the product of the output of each layer. The main motivation for this layered approach is the assumption that by proceeding this way we obtain two learning tasks that are simpler to solve than the original imbalanced task. The main reason for this simplification lies on the fact that the imbalance is strongly decreased on each of the tasks, thus making the modelling task simpler for most learning algorithms. Finally, we also remark that the proposed method automates the problem of defining the layers within the layered learning approach using a hierarchical clustering analysis. In previous works (e.g. \cite{cerqueira2020early}) this was carried out manually.

%In effect, the probability that a given instance belongs to the positive class is estimated according to the probability that it belongs to either the mixed or pure minority group times the probability that, given that the observations it belongs to either the mixed or pure minority group, it belongs to the positive class.

We carried out experiments using 100 benchmark binary classification data sets, and compared the proposed approach with several state-of-the-art methods. These include several resampling approaches, such as SMOTE \cite{chawla2002smote} and ADASYN \cite{he2008adasyn}, and a special-purpose algorithm designed for imbalanced problems, namely the balanced random forest \cite{chen2004using}.
The results suggest that \texttt{ICLL} outperforms other approaches significantly according to the area under the ROC curve (AUC) metric.

In summary, the main contributions of this paper are two-fold:
\begin{itemize}
    \item A novel method called \texttt{ICLL} for tackling binary imbalanced classification problems based on layered learning which does not require any parameters;
    \item An automatic framework for defining the layers in a layered learning methodology designed for classification.
\end{itemize}

The experiments and proposed method are publicly available\footnote{\url{https://github.com/vcerqueira/icll}}. The paper is organized as follows. 
In the next section we overview the literature related to our work. We focus on two specific topics: imbalanced classification and layered learning.
In Section \ref{sec:methodology}, we define the predictive task, and formalize the proposed approach which is named \texttt{ICLL}.
We provide empirical evidence for the predictive performance of our method in Section \ref{sec:experiments}, which includes a significance analysis based on the Bayes signed-rank test \cite{benavoli2017time}.
The results from our work are discussed in Section \ref{sec:discussion}, where we highlight the advantages of our approach and list its known limitations. Finally, we conclude the paper in Section \ref{sec:conclusions}.

\section{Related Work}\label{sec:related_work}

In this section we overview the literature related to our work. Section \ref{sec:rw_ic} lists the main approaches used in the literature that deal with imbalanced classification problems. Section \ref{sec:rw_ll} describes layered learning approaches.

\subsection{Imbalanced Classification}\label{sec:rw_ic}

Some of the most popular solutions used to tackle  class imbalance in classification problems are resampling methods. 
These approaches transform the training set to enhance the prevalence of the minority class. This involves a strategy based on under-sampling the majority class, over-sampling the minority class, or both. Since the resampling occurs before model fitting, these methods are agnostic to the learning algorithm.

The simplest resampling methods are random under-sampling (\texttt{RU}) and random over-sampling (\texttt{RO}).
\texttt{RU} randomly selects instances from the majority class, and removes those instances from the training set. \texttt{RO} also selects instances at random, but from the minority class. The selected instances are replicated in the training data.

\texttt{SMOTE} (Synthetic Minority Over-sampling Technique) is a widely used over-sampling approach. It works by synthesising new examples based on existing ones. This is accomplished by interpolating between instances from the minority class within their neighbourhood. More precisely, an observation from the minority class is selected at random, along with its $k$ nearest neighbours. Then, a new synthetic instance is created by interpolating between one of the $k$ neighbours and the selected observation. This process can be carried out several times, for example until the distribution of classes is balanced.
After the initial publication \cite{chawla2002smote}, several extensions of this method have been published (e.g. \cite{fernandez2018smote}).

\texttt{ADASYN} (Adaptive Synthetic) \cite{he2008adasyn} is another over-sampling method which follows a similar approach to \texttt{SMOTE} and creates new synthetic instances. The key difference is that \texttt{ADASYN} focuses on instances which are difficult to learn, i.e., close to the decision boundary.

Informed under-sampling of the majority class is also a common approach to deal with the imbalance problem that embeds some domain information in the selection of the majority class examples to be removed. One example of this approach is the One-Sided Selection (\texttt{OSS}) method \cite{kubat1997addressing}. \texttt{OSS} identifies and removes instances close to the decision boundary using Tomek links \cite{tomek1976two}.
Let $x_1$ and $x_2$ denote two instances from different classes, and $d(x_1, x_2)$ the distance between these examples. The pair ($x_1$,$x_2$) is considered a Tomek link if there exists no other instance with a smaller distance $d$ to either $x_1$ or $x_2$. Typically, these instances are considered to be noise or close to the decision boundary.

\texttt{Near-Miss} is another informed under-sampling method \cite{mani2003knn}. The main idea behind it is, contrary to the approach taken by \texttt{OSS}, it tries to retain only the instances from the majority class which are close to the decision boundary, instead of the other way around.

Resampling methods can be embedded in some learning algorithms. For example, the balanced random forest \cite{chen2004using} extends the original method \cite{breiman2001random} by applying random under-sampling in each bootstrap sample. Several ensemble methods have been modified through the integration of resampling to tackle the class imbalance problem~\cite{galar2011review}. For instance, \texttt{SMOTEBoost}~\cite{chawla2003smoteboost} and \texttt{SMOTEBagging}~\cite{wang2009diversity} integrate \texttt{SMOTE} into a boosting and bagging ensemble, respectively, There are also hybrid ensembles that combine both bagging and boosting with resampling methods (e.g. \texttt{EasyEnsemble} and \texttt{BalanceCascade}~\cite{liu2008exploratory}).

Previous works in the literature tried to couple clustering with resampling methods, e.g. \cite{wu2007local,nickerson2001using,bellinger2019cure}. For example, \texttt{CURE} (Clustered Resampling) is a resampling approach which applies clustering analysis to the data and can perform both over-sampling and under-sampling \cite{bellinger2019cure}. First, the method learns the structure of the data using hierarchical clustering analysis. 
Their approach for hierarchical clustering includes a novel semi-supervised metric used to compute the distance between each instance, and a novel heuristic for cutting the hierarchy and forming the clusters.
Under-sampling of instances from the majority class is only performed in cases not close to the decision boundary. On the other hand, over-sampling of minority instances is carried out with instances from the same concept or cluster.
Our proposed method leverages ideas from  \texttt{CURE}. In particular, as we will detail in the next section, we cluster the observations using a hierarchical method. We also apply the heuristic developed by the authors of \texttt{CURE} to cut the hierarchy and form the clusters.
However, while \texttt{CURE} applies a semi-supervised mechanism for clustering our approach is purely unsupervised and based on the Euclidean distance. Finally, the purpose of our clustering analysis is to automatically create the layers for the layered learning methodology rather than to resample the training data.

The proposed approach is automated insofar as it does not require any parameters. In this context, it can be regarded as an automated machine learning (AutoML) approach for imbalanced problems, specifically binary classification. AutoML is becoming increasingly relevant in the machine learning literature. In the case of imbalanced problems, the literature is scarce. Moniz et al. \cite{moniz2021automated} presented a recent work which leverages meta-learning to select the best resampling strategy to apply in a given data set. Li et al. \cite{li2021autobalance} developed a automatic loss function for training deep neural networks.

\subsection{Layered Learning}\label{sec:rw_ll}

Layered learning denotes a hierarchical procedure in which a predictive task is decomposed into simpler sub-tasks or layers, and each layer influences the learning process of the subsequent ones. For example, this may occur by influencing which training instances or predictor variables are used.

Layered learning was originally introduced by Stone and Veloso \cite{stone2000layered} for robotic soccer, where they split the task of \textit{passing a ball} into three sub-tasks: (i) intercepting the ball; (ii) evaluation of passing possibilities; and (iii) sending the ball.
More generally, splitting a task into different parts is a common approach in the hierarchical reinforcement learning literature, such as the options framework devised by Sutton et al. \cite{sutton1999between}.

Cerqueira et al. \cite{cerqueira2020early} have applied a layered learning process to tackle classification problems, specifically the detection of impending critical health episodes in the intensive care unit of hospitals. They, and subsequent related works \cite{ribeiro2021layered}, show the advantage of this approach relative to a standard classification strategy coupled with resampling pre-processing methods.
The crucial difference to our work is that they manually define the layers according to domain expertise or by optimization. Conversely, we provide a novel approach which accomplishes this task automatically using hierarchical clustering analysis. Moreover, we also systematise layered learning approaches for imbalanced binary classification problems.

There are similar hierarchical procedures to layered learning in the literature. For example, cascade classifiers~\cite{sharma2012anomaly} denote a stepwise methodology in which a meta-classifier is built to predict which concept an instance refers to. Then that instance is passed to the appropriate base model. The critical difference of this approach to layered learning is that the learning of each layer affects the learning of subsequent ones. Besides, each instance traverses all layers instead of being routed to a specific model.

\section{Methodology}\label{sec:methodology}

In this section we start by defining the predictive task addressed in this work (Section \ref{sec:problem}). Next we formalize the method we propose to tackle this task (Section \ref{sec:method}). 

\subsection{Problem Definition}\label{sec:problem}

Let $\mathcal{D}$ denote a data set, which is defined as $\mathcal{D} = \{\langle x_i, y_i \rangle\}^n_{i=1}$, where $x_i \in X$ represents the feature vector for the $i$-th instance, and $y_i \in Y$ represents the respective target value. The target variable $Y$ is discrete and can take two values.
The objective is to carry out supervised learning, which, given the domain of $Y$, amounts to a binary classification task.
Moreover, let $\mathcal{D}_{0}$ denote a subset of $\mathcal{D}$ in which all $y \in \mathcal{D}_{0}$ belong to the class 0; $\mathcal{D}_{1}$ represents the subset of observations for the class 1. An imbalanced classification problem arises because the number of observations belonging to one class is significantly larger than the number of observations of the other class: $|\mathcal{D}_{0}| >> |\mathcal{D}_{1}|$. We will refer to 0 as the majority class, and 1 to the minority class.

\subsection{Imbalanced Classification Layered Learning}\label{sec:method}

In this sub-section we formalize the proposed method which is named \texttt{ICLL} for Imbalanced Classification Layered Learning.
The training of the \texttt{ICLL} method is based on three main steps:

\begin{enumerate}
    \item Clustering the training instances using a hierarchical clustering method \cite{murtagh2011methods};
    
    \item Automatic creation of the layers for the layered learning strategy based on the output of the clustering model;
    
    \item Training a predictive model in each layer.
\end{enumerate}

\noindent In the rest of this section we will describe each step in detail.

\subsubsection{Step 1: Clustering Analysis}

In the first stage of the methodology a hierarchical clustering algorithm is applied to cluster the training instances. The main motivation for clustering is to extract structural information from the data, which will be used to automatically define the layers.
Hierarchical clustering methods group data into a tree structure, which provides different levels of abstraction of that data \cite{murtagh2011methods}.
We adopt an agglomerative approach for the clustering algorithm, which is a common strategy that can be described as follows.
Each instance is first assigned as their own cluster. Then, pairs of clusters are successively merged until there is only a single cluster that contains all observations. 
One of the advantages of using a hierarchical algorithm for clustering is that it does not require the number of clusters as an input parameter.

\IncMargin{1em}
\begin{algorithm}[ht]
    \SetKwInOut{Input}{Input}
    \SetKwInOut{Output}{Output}

    \Input{X - Predictor variables for classification training data set}
    \Output{$C$ - Set of clusters}

    \BlankLine
    
    $M_X \leftarrow $ PairwiseDistance($X$, method $=$ Euclidean) \tcp{Pairwise distance matrix of $X$ using the Euclidean distance}
    
    $Z \leftarrow $ Linkage($M_X$, method=Ward) \tcp{Agglomerative linkage tree using $M_X$ and the Ward method}
    
    InterClusterDistance(Z) $\leftarrow $ log(InterClusterDistance(Z)) \tcp{Log transformation of the inter-cluster distances obtained in the tree $Z$}
    
    $\mu \leftarrow $ Mean(InterClusterDistance(Z)) \tcp{Mean of the inter-cluster distances}
    
    $\sigma \leftarrow $ StandardDeviation(InterClusterDistance(Z)) \tcp{Standard deviation of the inter-cluster distances}
    
    $\tau \leftarrow \mu + \sigma$ \tcp{Maximum inter-cluster distance for cluster formation}
    
    $C \leftarrow $ FormClusters(Z, threshold=$\tau$) \tcp{Form clusters using the linkage tree, subject to the threshold $\tau$}
    
    Return $C$
\caption{Hierarchical clustering analysis}
\label{alg:step1}
\end{algorithm}
\DecMargin{1em}

The hierarchical clustering process is detailed in Algorithm \ref{alg:step1}.
In order to group the data we start by measuring the dissimilarity between each pair of instances according to the Euclidean distance (line 1). As the linkage criterion, we apply the Ward method, which minimizes the total within-cluster variance (line 2). We use the Ward method to bias the clusters to contain instances that are highly concentrated. In other words, the goal is to obtain clusters which represent sub-concepts or groups of highly similar instances. These groups will then be sorted according to those that are easy and hard to classify (in the next step of the methodology).

%Regarding the stopping criterion for the clustering algorithm, and subsequent cluster formation, we follow the strategy developed by Bellinger et al. \cite{bellinger2019cure} (lines 3--7). Essentially, each instance is assigned to the cluster with the largest cardinality that it belongs to and has an intra-cluster distance less than the threshold $\tau = \mu + \sigma$; $\mu$ and $\sigma$ denote the mean and standard deviation of intra-cluster distances. 

% Colin's comment 
Once the cluster hierarchy is formed, we must extract a specific clustering of the data from the hierarchy. Bellinger et al. \cite{bellinger2019cure} proposed an automatic strategy by which to extract this clustering such that it accounts for the natural spread in the data. The objective is to define clusters that are large enough to capture entire sub-concepts without including multiple sub-concepts. In this method (lines 3--7), each instance is assigned to the cluster with the largest cardinality that it belongs to and has an intra-cluster distance less than the threshold $\tau = \mu + \sigma$; $\mu$ and $\sigma$ denote the mean and standard deviation of the log transformed intra-cluster distances. Intuitively, utilising $\mu + \sigma$ as the threshold caps distance between the samples in the clusters at a level that is natural according the target dataset. The log transformation is applied because Bellinger et al. \cite{bellinger2019cure} discovered that the intra-cluster distances approximately followed a lognormal distribution.

\subsubsection{Step 2: Automatic Layer Definition}\label{sec:step2}

\begin{figure}[t]
    \centering
    \includegraphics[width=\textwidth, trim=0cm 0cm 0cm 0cm, clip=TRUE]{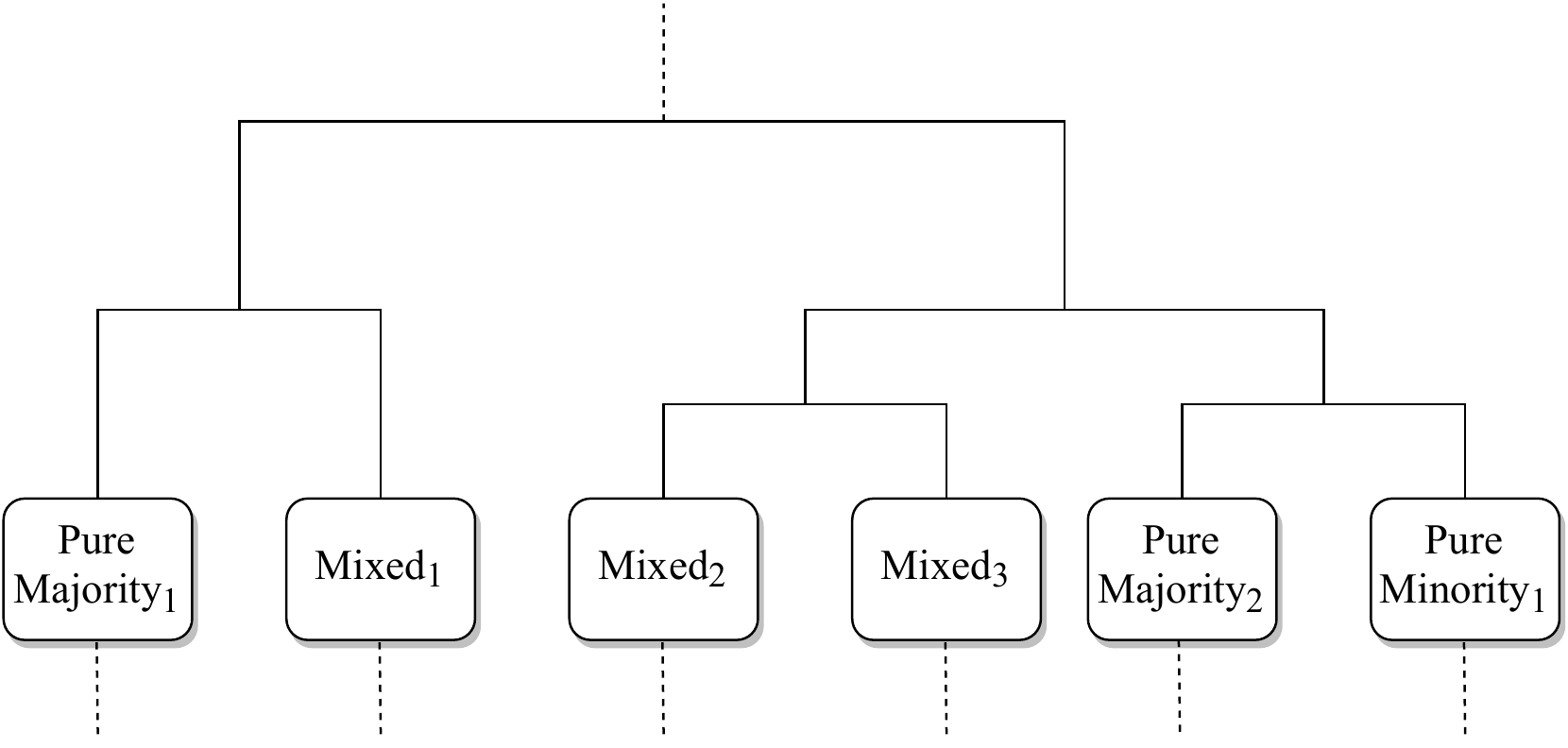}
    \caption{An example of a partial dendrogram resulting from the hierarchical clustering. The instances are assigned to three possible groups according to the class distribution of the respective cluster after the cut-off.}
    \label{fig:example}
\end{figure}

At this point, we have the cluster allocation for each training observation $(x, y) \in \mathcal{D}$.
In the second stage of the methodology we create the layers which comprise the layered learning strategy. We follow the layer architecture developed by Cerqueira et al. \cite{cerqueira2020early} for early event detection and construct a workflow with two layers.

\IncMargin{1em}
\begin{algorithm}[ht]
    \SetKwInOut{Input}{Input}
    \SetKwInOut{secInput}{    }
    \SetKwInOut{Output}{Output}

    \Input{$\mathcal{D}$ - Binary classification data set}
    \secInput{$C$ - Set of clusters}
    \Output{$C_{maj}, C_{min}, C_{mix}$}

    \BlankLine
    
    $C_{maj} \leftarrow $ $[$ $]$  
    \tcp{Initialize empty group of pure majority instances}
    
    $C_{min} \leftarrow $ $[$ $]$  
    \tcp{Initialize empty group of pure minority instances}
    
    $C_{mix} \leftarrow $ $[$ $]$  
    \tcp{Initialize empty group of borderline instances}
    
    \ForEach{\text{instance} ($x_i$, $y_i$) in $\mathcal{D}$}{
    
    $c_i \leftarrow $ Cluster of ($x_i$, $y_i$) 
    
    \uIf{$c_i$ \text{contains only instances of the majority class}}{
        $C_{maj}.add$(($x_i$, $y_i$)) \; \tcp{Add instance to $C_{maj}$}
    }
    \uElseIf{$c_i$ \text{contains only instances of the minority class}}{
        $C_{min}.add$(($x_i$, $y_i$)) \;
        \tcp{Add instance to $C_{min}$}
    }
    \Else{
        $C_{mix}.add$(($x_i$, $y_i$)) \; \tcp{$c_i$ contains instances from both classes -- add instance to $C_{mix}$}
        
    }
    
    }
    
    Return $C_{maj}, C_{min}, C_{mix}$
\caption{Group assignment based on cluster class distribution}
\label{alg:step2}
\end{algorithm}
\DecMargin{1em}

The two layers are automatically defined according to the clustering composition obtained in the previous step.
Specifically, we assign each instance to one of three possible groups:
\begin{itemize}
    \item Pure majority group: if the corresponding cluster comprises only instances of the majority class;
    \item Pure minority group: if the corresponding cluster comprises only instances of the minority class;
    \item Mixed group: if the corresponding cluster contains observations from both majority and minority classes.
\end{itemize}

\noindent This idea is depicted in Figure \ref{fig:example}, where part of a dendrogram is shown. In this synthetic example there are six clusters, two of each comprise only majority instances (Pure Majority 1 and 2), one contains only instances of the minority class (Pure Minority 1), and the remaining ones are Mixed.
The process of assigning the observations into the three possible groups is formalized in Algorithm \ref{alg:step2}.

This process enables the understanding of the position of each training instance with respect to the decision boundary. If an instance belongs to the pure majority group it can be said that it is clear that this observation belongs to the majority class. In other words, the instance is arbitrarily far from the decision boundary on the side of the majority class. The same can be argued for instances belonging to the pure minority group. In principle, these observations are far from the decision boundary but on the side of the minority class.
Finally, we have instances from the mixed group whose cluster comprises instances from both minority and majority classes. We regard these observations as borderline instances, i.e. arbitrarily close to the decision boundary.

In order to leverage the information regarding the group of each training instance we adopt a stepwise methodology based on layered learning.
As we described in Section \ref{sec:rw_ll}, layered learning denotes a learning approach in which a predictive task is split into multiple ones which are, in principle, easier to solve.
We construct two layers, Layer 1 and Layer 2, according to the definitions below. 

\paragraph{Layer 1}

The two layers denote two sub-tasks of the original problem. These sub-tasks comprise a more balanced class distribution. We hypothesise that this will lead to easier classification tasks and, consequently, better performance.
Let L1 denote the event ``The observation belongs to the Mixed or Pure Minority group".
For the first layer, the target value for a given instance is defined as:

\begin{equation}\label{eq:task1b}
  y^{L1}_i = \begin{cases}
            1 & \text{if L1 happens},\\
            0 & \text{otherwise}.
          \end{cases}
\end{equation}

Essentially, we attempt to model whether or not each observation belongs to a purely majority group, in which case $y^{L1} = 0$, or not ($y^{L1} = 1$). In practice, the original predictor variables remain the same, but the target values $y \in Y$ are replaced with $y^{L1} \in Y^{L1}$, where $Y^{L1}$ denotes the target variable for the layer L1.

This first layer can be regarded as an approach designed to distinguish instances from the majority class which are \textit{easy} to learn (belong to the pure majority group) from the others, which are either borderline (mixed group) or scarce (pure minority group). 
 
\paragraph{Layer 2}

Assuming that the first layer is modelled successfully, the second layer attempts to solve the remaining problem: Given that L1 occurs, i.e., an observations belongs to the mixed or pure minority group, we want to find whether it belongs to the minority class (i.e. $y_i = 1$). The target variable for this sub-task ($y^{L2}$) is formalised in Equation \ref{eq:task2}.

\begin{equation}\label{eq:task2}
 \text{Given }y^{L1}\text{ = 1, } y^{L2}_{i} = \begin{cases}
            1 & \text{if } y_i = 1,\\
            0 & \text{otherwise}.
          \end{cases}
\end{equation}

In the second layer we discard all observations from the pure majority group, and only fit a model with those from the mixed and pure minority group ($y^{L1} = 1$ $\forall y^{L1} \in Y^{L1}$). %This instance subset denotes the main process by which the first layer influences the second layer.

\subsubsection{Step 3: Model Fitting and Inference}

% discuss error propagation problem?

In the training stage, a predictive model is fit in each one of the two layers. We note that, since the two layers are independent, the training can occur in parallel.

Let $f^{L1}$ and $f^{L2}$ denote the models trained in the layers L1 and L2, respectively. We combine the output of these models according to a function $g$:

\begin{equation}\label{eq:merget1t2}
    g(x_i) = f^{L1}(x_i) \cdot f^{L2}(x_i)
\end{equation}

\noindent The final decision is made according to the multiplication of the individual predictions. Therefore, our approach postulates that the probability that a given instance belong to the minority class is estimated according to the probability that it belongs to either the mixed or pure minority group times the probability that, given that the observations it belongs to either the mixed or pure minority group, it belongs to the minority class. This process is also applicable to non-probabilistic classifiers. That is, our approach predict that a given instance $x_i$ belongs to the minority class if both $f^{L1}(x_i)$ and $f^{L2}(x_i)$ predict 1.

\subsection{Methodological Limitations}\label{sec:limitations}

The layer definition stage of the methodology strongly depends on the outcome of the hierarchical clustering model. We formalized our method assuming that all three groups $C_{maj}$, $C_{min}$, and $C_{mix}$ are non-empty, but this might not be the case depending on the input data.

The methodology works normally in the case that $C_{min}$ is empty, i.e. there are no instances in the pure minority group. By definition, the minority class comprises a relatively low number of observations. In effect, it is expected that, even if the minority class is more prevalent in a given cluster, all clusters contain some instances from the majority class.

If $C_{mix}$ is empty, this means that the hierarchical clustering is able to perfectly split the two classes in the chosen cut-off. In this case the proposed methodology becomes redundant. However, this also means that the predictive task is solved.

The scenario in which $C_{maj}$ is empty is more problematic as the first layer cannot be defined properly. Our approach to solve this problem would be to change the threshold $\tau$ and move the formation of the clusters up in the hierarchy.
Notwithstanding we remark that, given the relative high prevalence of instances of the majority class, this represents a highly unlikely scenario.

\section{Experiments}\label{sec:experiments}

This section describes the experiments carried out to validate \texttt{ICLL}. These were designed to address the following research questions:

\begin{enumerate}
    \item \textbf{RQ1}: How does \texttt{ICLL} perform relatively to state-of-the-art methods for IBC problems?;
    \item \textbf{RQ2}: Does applying resampling to the proposed method improve its performance? Since resampling methods are agnostic to the learning algorithm, we attempt to couple these approaches with \texttt{ICLL} and assess whether performance improvements are obtained;
    \item \textbf{RQ3}: Is \texttt{ICLL} a model-based strategy for under-sampling? The second layer of \texttt{ICLL} is carried out after discarding observations which are arbitrarily far from the decision boundary on the side of the majority class. Thus, it can be argued that the first layer is performing under-sampling and the discrimination between classes occurs in the second layer. We will test this hypothesis in the experiments;
    \item \textbf{RQ4}: Are the conclusions consistent when taking into account only \textit{difficult} problems? We examine the results of the experiments with all available data sets (presented in Section \ref{sec:casestudy}) and using only a subset were a baseline performs poorly.
\end{enumerate}

\subsection{Case Study and Experimental Design}\label{sec:casestudy}

\begin{figure}[h]
    \centering
    \includegraphics[width=\textwidth, trim=0cm 0cm 0cm 0cm, clip=TRUE]{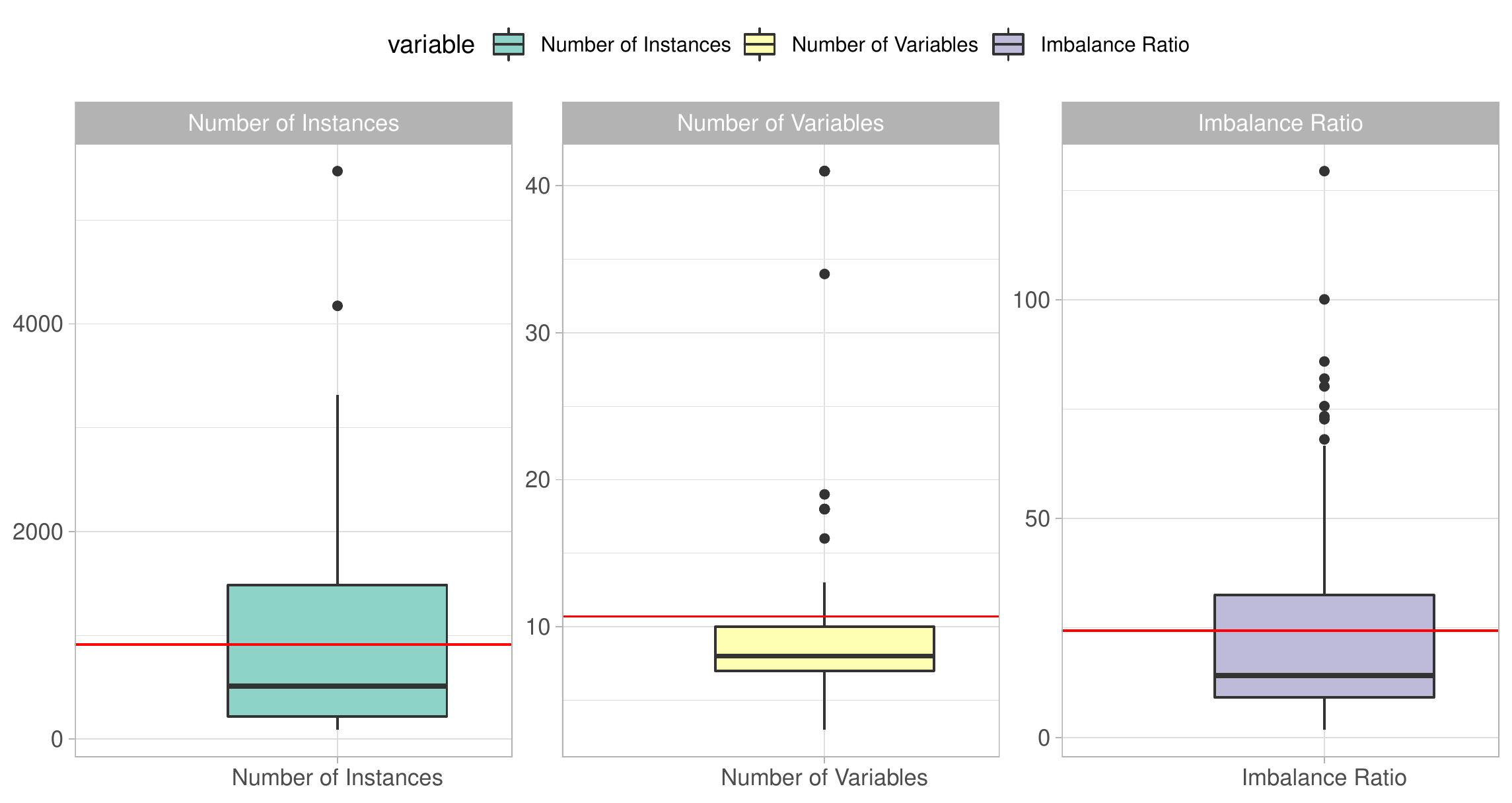}
    \caption{Boxplots showing the distribution of number of instances, number of explanatory variables, and imbalance ratio of the data sets used in the experiments. The horizontal red line denotes the mean value for each characteristic.}
    \label{fig:dataset}
\end{figure}

The experiments where carried out using 100 data sets. These were retrieved from the KEEL repository \cite{fernandez2008study,fernandez2009hierarchical,KEEL}, which provides benchmark data sets for imbalanced domain learning. From this repository, we collected all binary imbalanced classification data sets. The distribution of basic characteristics of these 100 data sets are shown in Figure \ref{fig:dataset}, namely the number of observations, number of explanatory variables, and imbalance ratio. The number of instances range from 92 to 5472 with an average of 934 data points, while the number of variables range from 3 to 100 with an average of 11.6. The minimum, maximum, and average imbalance ratio is 1.8, 129.4, and 25.7 respectively. In effect, the case study covers data sets with different imbalance ratio profiles.

We applied a 2$\times$5-fold stratified cross-validation procedure for estimating the predictive performance of each approach. The performance was measured according to the area under the ROC curve (AUC). 
In terms of learning algorithms, we tested a Random Forest (RF), a Support Vector Machine (SVM), and a Logistic Regression (LR). We resorted to the implementation from scikit-learn \cite{pedregosa2011scikit} to apply these algorithms with their default parameters. The RF provided the overall best results. In effect, we will show the complete results only for this method. Notwithstanding, we include a variant of the remaining approaches in the interest of completeness. All conclusions also hold for these methods. We also remark that the Random Forest is used in both layers of \texttt{ICLL}, but this is not a requirement of the method. Different learning algorithms could be applied in each layer.

\subsection{Methods}

Besides the proposed approach \texttt{ICLL}, we include the following methods in the experiments:

\begin{itemize}
    \item \texttt{NoResample-RF}, \texttt{NoResample-SVM}, \texttt{NoResample-LR}: A standard binary classifier which is fit in the training data set without any specific mechanism for dealing with the imbalanced class distribution. As explained above we test three learning algorithms for this approach;
    
    \item \texttt{SMOTE}, \texttt{CURE}, \texttt{ADASYN}, \texttt{NearMiss}, \texttt{OSS}, \texttt{RO}, \texttt{RU}: A Random Forest classifier is fit after the training data is pre-processed with the respective resampling method for balancing the class distribution. These approaches were described in Section \ref{sec:rw_ic};
    
    \item \texttt{BalancedRF}: A variant of the Random Forest algorithm in which random under-sampling is carried out for each bootstrap sample \cite{chen2004using};
    
\end{itemize}

\noindent We remark that all resampling approaches were applied using their default parameter setting according to the implementation provided by the \textit{imblearn} python library\footnote{\url{https://pypi.org/project/imblearn/}}. Besides these state-of-the-art approaches, we also include the following five variants of \texttt{ICLL}:

\begin{itemize}
    \item \texttt{ICLL+SMOTE}: The proposed method described in Section \ref{sec:method} works without any resampling approach. In this variant we couple our approach with a SMOTE \cite{chawla2002smote} resampling method. Essentially, we apply SMOTE in each of the two layers to balance the distribution of the classes;
    
    \item \texttt{ICLL+SMOTE(L1)}: This approach is similar to \texttt{ICLL+SMOTE}, but \texttt{SMOTE} is applied only in the first layer;
    
    \item \texttt{ICLL+SMOTE(L2)}: Another approach similar to \texttt{ICLL+SMOTE}, in which \texttt{SMOTE} is applied only in the second layer;
    
    \item \texttt{ICLL(L2)}: A variant which only uses the output from the second layer. It can be argued that the proposed layered learning method is performing a model-based under-sampling in the first layer. Thus, the performance advantage is obtained only in the second layer, making the first layer unnecessary in the inference stage. We test this hypothesis using this variant.
    
    \item \texttt{ICLL(L1)}: In the interest of completeness, we also include a variant which uses only the output from the first layer. 
\end{itemize}

\subsection{Results}

We present the results of the experiments in this section. First (Section \ref{sec:pt1}), we perform a preliminary analysis on the applicability of the proposed method. Then (Section \ref{sec:pt2}), we compare all methods according to the average rank and measure the percentage difference in predictive performance. Finally, we repeat this analysis but only taking into account the datasets where the approach \texttt{NoResample-RF} does not provide a good predictive performance (Section \ref{sec:pt3}).

\subsubsection{Preliminary Analysis}\label{sec:pt1}

In this section we present the results of the experiments. As we explained in Section \ref{sec:limitations} there are scenarios in which the stepwise approach followed by \texttt{ICLL} becomes redundant, specifically when the mixed group ($C_{mix}$) is empty. As we mentioned, this means that the hierarchical clustering method was able to perfectly split the majority instances from the minority instances to different clusters. This case occurred in at least one of the ten iterations of 11 (out of 100) data sets. 
%When this occurs, we output the predictions made by the method \texttt{NoResample-RF}, which does not apply any special mechanism to deal with the class imbalance problem. Our rationale is that, if the clustering model is able to split the classes perfectly, then a standard classifier should be able to do so as well. 
In principle, if the clustering model is able to split the classes perfectly then a standard classifier should be able to do so as well. Indeed, in all 11 datasets where this issue occurred, the final AUC score of the \texttt{NoResample-RF} approach was 1 (i.e. a perfect score). 
We continue the analysis without these 11 datasets. 

\subsubsection{Average Rank, Magnitude of Differences, and Significance Analysis}\label{sec:pt2}

\begin{figure}[h]
    \centering
    \includegraphics[width=\textwidth, trim=0cm 0cm 0cm 1cm, clip=TRUE]{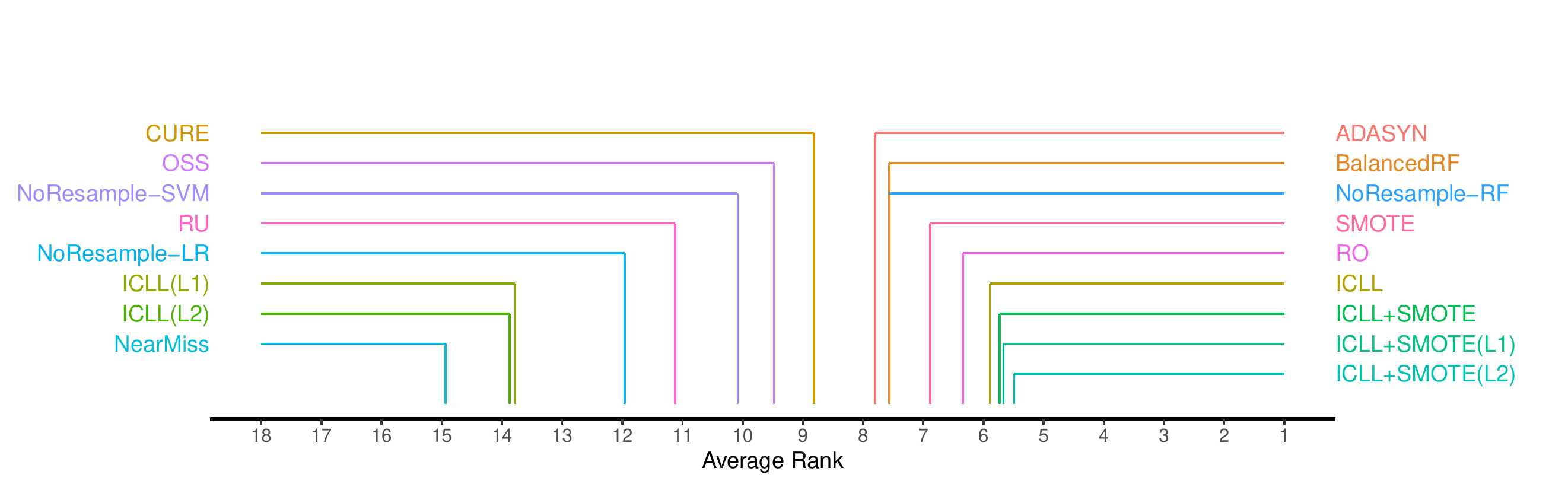}
    \caption{Average rank of each method. Lower values denote better performance.}
    \label{fig:avgrank}
\end{figure}

Regarding the analysis of results we start by observing the average rank of each method across the remaining 89 problems. Afterwards, we analyse the magnitude in the differences in performance to assess their significance. In Figure \ref{fig:avgrank} we show a diagram that depicts the average rank of each method across the 89 data sets. A given method has a rank of 1 in a given problem if it shows the best performance (AUC) in that problem. Effectively, the average rank represents the average position of each approach relative to the remaining ones.

Four of the variants of \texttt{ICLL} are the best four methods in average rank. The best one is \texttt{ICLL+SMOTE(L2)}, which applies the SMOTE resampling method in the second layer. The version of the proposed method which does not apply any resampling approach (\texttt{ICLL}) shows a better score than any state-of-the-art resampling approach. Notwithstanding, its performance improves when coupled with SMOTE.
The variants of the proposed method which only use the output of one of the layers (\texttt{ICLL(L1)} and \texttt{ICLL(L2)}) show one of the worse average ranks. This suggests that the output of both layers is critical for the predictive performance of \texttt{ICLL}.
Regarding state-of-the-art resampling methods, \texttt{RO} shows the best average rank, followed by \texttt{SMOTE}.
Finally, the standard classifier trained with a Random Forest (\texttt{NoResampler-RF}) shows a better score relative to the same approach trained with either a Logistic Regression (\texttt{NoResampler-LR}) or a SVM (\texttt{NoResampler-SVM}).

\begin{figure}[h]
    \centering
    \includegraphics[width=\textwidth, trim=0cm 0cm 0cm 0cm, clip=TRUE]{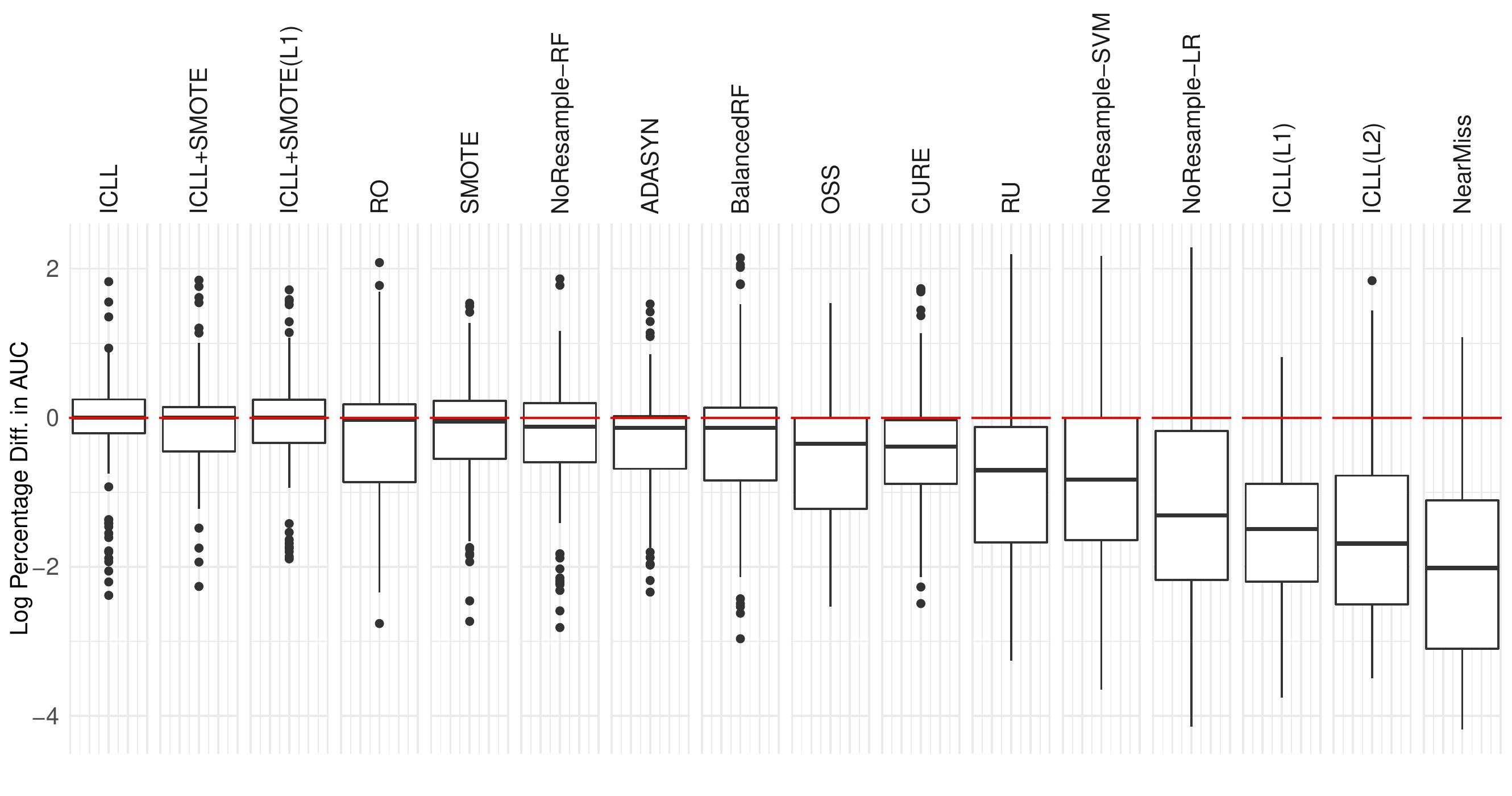}
    \caption{Boxplots showing the distribution of the log percentage difference between each method and \texttt{ICLL+SMOTE(L2)}. Negative values denote better performance of \texttt{ICLL+SMOTE(L2)}.}
    \label{fig:bp_pd_all}
\end{figure}

As mentioned before, the average rank measures the average position of each method relative to the remaining ones. However, it does not take into account the magnitude of differences of predictive performance. Therefore, we complement the average rank by analysing the percentage difference in performance between each method and the variant of the proposed approach with best average rank (\texttt{ICLL+SMOTE(L2)}). This can be formalized as follows for a given method \texttt{m}:

\begin{equation*}
    100 \times \frac{\text{AUC}_{\texttt{m}} - \text{AUC}_{\texttt{ICLL+SMOTE(L2)}} }{\text{AUC}_{\texttt{ICLL+SMOTE(L2)}}}
\end{equation*}

\noindent where $\text{AUC}_{\texttt{m}}$ and $\text{AUC}_{\texttt{ICLL+SMOTE(L2)}}$ represent the AUC of method \texttt{m} and \texttt{ICLL+SMOTE(L2)}, respectively. Since AUC should be maximized, negative values in percentage difference denote better performance of \texttt{ICLL+SMOTE(L2)}.

We show this analysis in Figure \ref{fig:bp_pd_all}. This figure depicts several boxplots showing the distribution of the log percentage difference between each method and \texttt{ICLL+SMOTE(L2)}, where negative values denote better performance for the proposed method. Moreover, the methods are ordered by decreasing median percentage difference in AUC. Thus, more competitive methods appear first (from left to right).
In terms of ranking, the order of the methods is similar to that obtained according to the average rank analysis. The main take away is that, for all methods, most of the distribution lies below the zero line. This shows that \texttt{ICLL+SMOTE(L2)} outperforms the other methods more times than not.

While it is clear that \texttt{ICLL+SMOTE(L2)} shows a better performance, Figure \ref{fig:bp_pd_all} also shows that the percentage difference is close to zero in many cases. In this context, we performance a new analysis which considers small differences in performance to be negligible and the pair of models practically equivalent. 

\begin{figure}[h]
    \centering
    \includegraphics[width=\textwidth, trim=0cm 0cm 0cm 0cm, clip=TRUE]{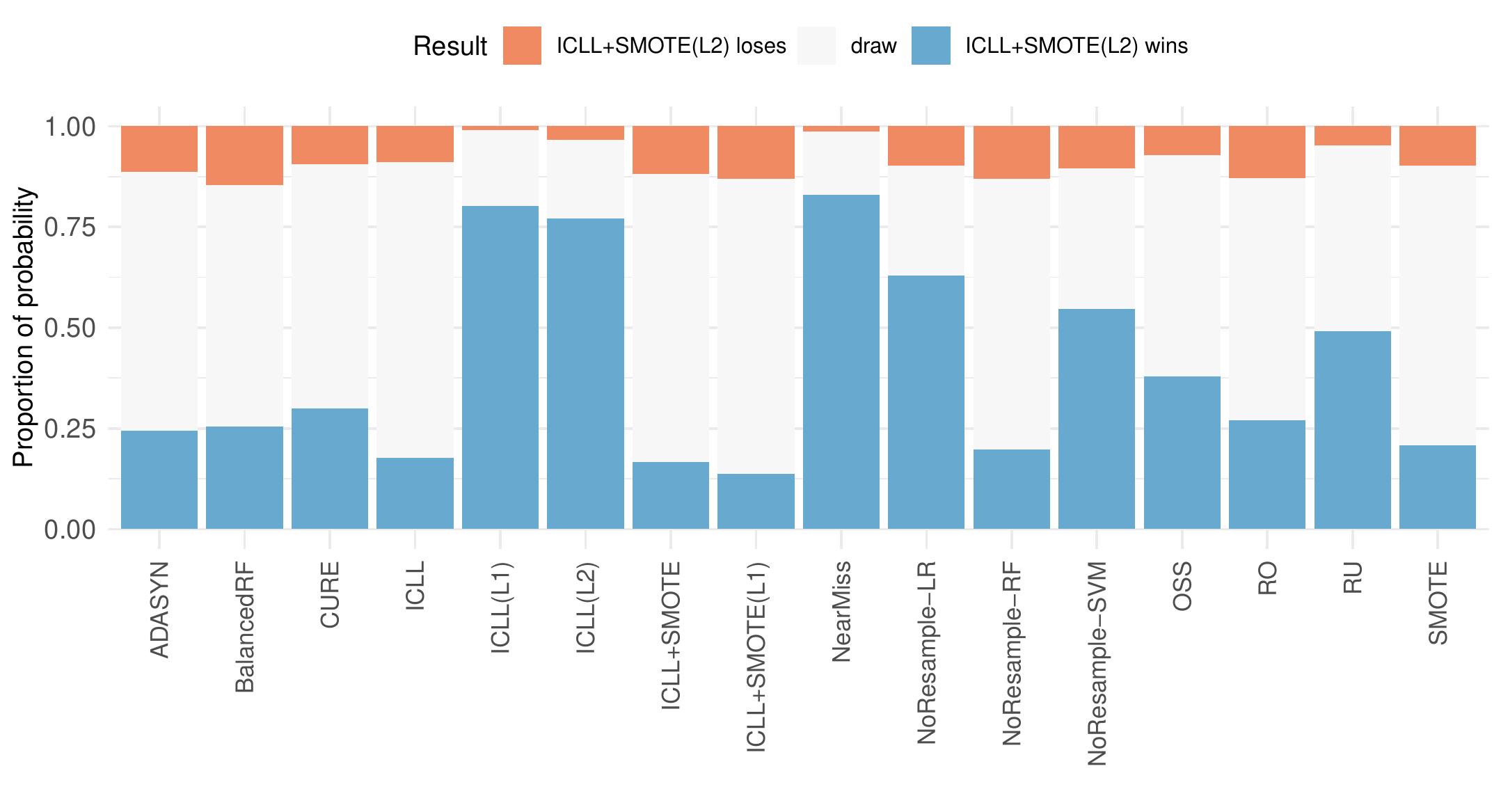}
    \caption{Paired comparisons between \texttt{ICLL+SMOTE(L2)} and each remaining approach. Each stacked barplot shows the probability of \texttt{ICLL+SMOTE(L2)} winning in blue (result below -1\%), drawing in light grey (result withing [-1\%, 1\%]), or losing in red (result above 1\%).}
    \label{fig:prob_ratios}
\end{figure}

We define this interval to be [-1\%, 1\%]. This means that the performance of two methods under comparison is considered equivalent if their percentage difference is within this interval.

Figure \ref{fig:prob_ratios} shows the probability of \texttt{ICLL+SMOTE(L2)} winning in blue (percentage difference below -1\%), drawing in light grey (results within [-1\%, 1\%]), or losing in red (percentage difference above 1\%) against each remaining method. For example, relative to \texttt{ADASYN}, \texttt{ICLL+SMOTE(L2)} has a probability of winning of around 25\%, a probability of losing of about 12.5\%, and a probability of drawing of about 62.5\%.
Analysing the scores, it is clear that \texttt{ICLL+SMOTE(L2)} outperforms the other methods. That is, the probability of winning is larger than the losing. Nonetheless, there is a considerable probability that the results end up being comparable (a percentage difference in AUC below 1\%).

\begin{figure}[h]
    \centering
    \includegraphics[width=\textwidth, trim=0cm 0cm 0cm 0cm, clip=TRUE]{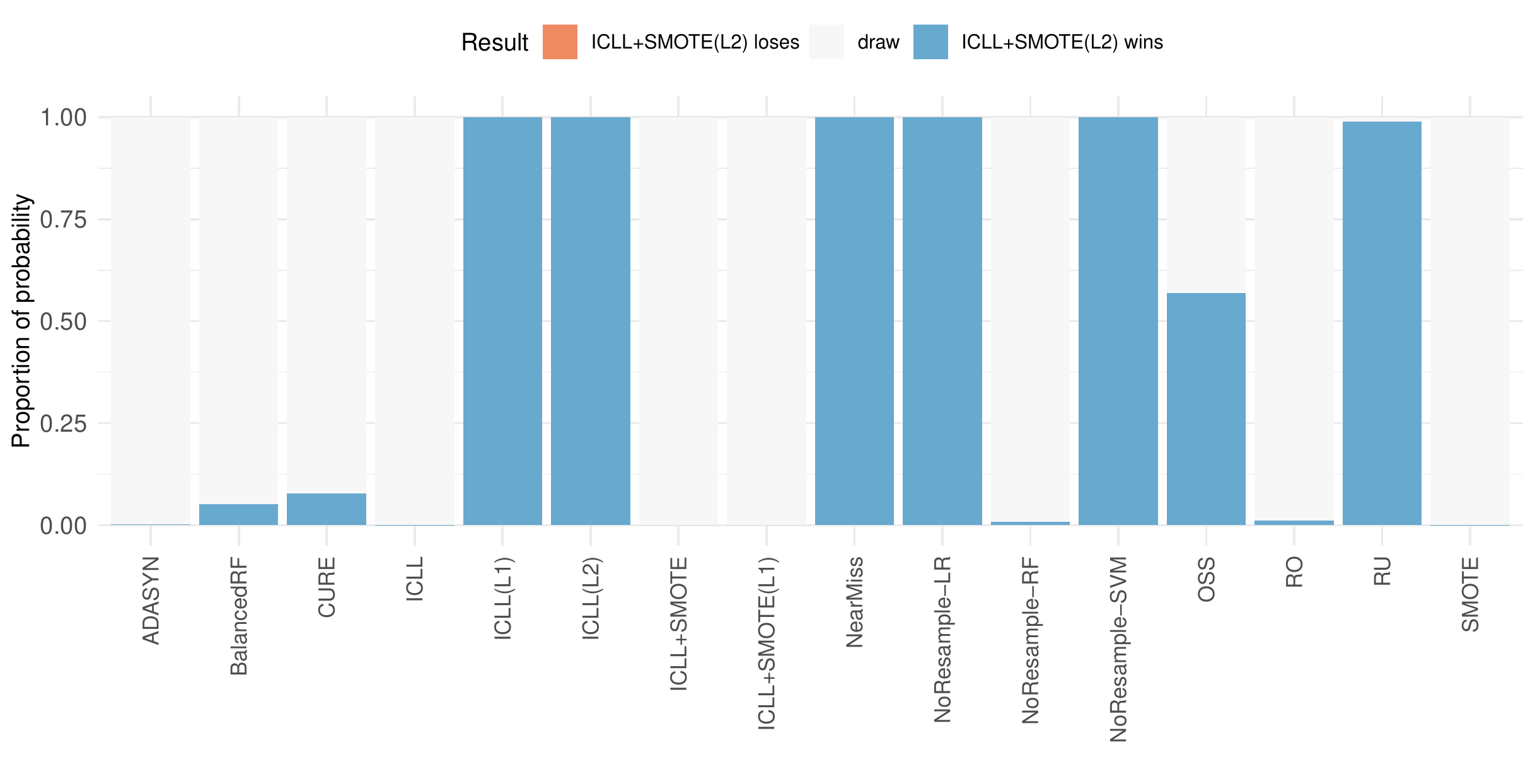}
    \caption{Paired comparisons between \texttt{ICLL+SMOTE(L2)} and each remaining approach using the Bayesian signed-rank test. Each bar represents the proportion of probability of each outcome: the blue part denotes the probability that \texttt{ICLL+SMOTE(L2)} wins significantly; the red part represents the proportion of probability in which \texttt{ICLL+SMOTE(L2)} loses significantly; the grey area represents the probability of a draw.}
    \label{fig:bayes_all}
\end{figure}

Finally, we carried out a bayesian analysis to assess the significance of the results using the Bayes signed-rank test \cite{benavoli2017time}. This test is used to compare pairs of predictive models across multiple data sets. In this case, we compare \texttt{ICLL+SMOTE(L2)} with all remaining methods.
We also define the region of practical equivalence (ROPE) for the Bayes signed-rank test to be the interval [-1\%, 1\%]. The results are shown in Figure \ref{fig:bayes_all}, which follow a similar structure as Figure \ref{fig:prob_ratios}. 
The results of the test show that \texttt{ICLL+SMOTE(L2)} either wins significantly with high probability or draws when compared with other methods and considering a 1\% ROPE level. While \texttt{ICLL+SMOTE(L2)} tends to outperform state-of-the-art approaches, the differences are often small and not statistically significant above 1\% when compared with approaches such as \texttt{NoResample-RF} or \texttt{SMOTE}.

\subsubsection{Repeating the Study for Difficult Problems}\label{sec:pt3}

In many of the data sets, specifically in 71 out of 100, the \texttt{NoResample-RF} approach is able to achieve at least 0.9 AUC. This means that, even without any special mechanism for dealing with the imbalanced distribution, the predictive model is able to distinguish between both classes with a good performance. 
In this context, we decided to repeat the result analysis but only considering the data sets in which \texttt{NoResample-RF} has an AUC score lower than 0.9. 
Overall, there are 29 data sets where this occurs. For simplicity, we will refer to these data sets as the difficult problems within our case study.

\begin{figure}[h]
    \centering
    \includegraphics[width=\textwidth, trim=0cm 0cm 0cm 0cm, clip=TRUE]{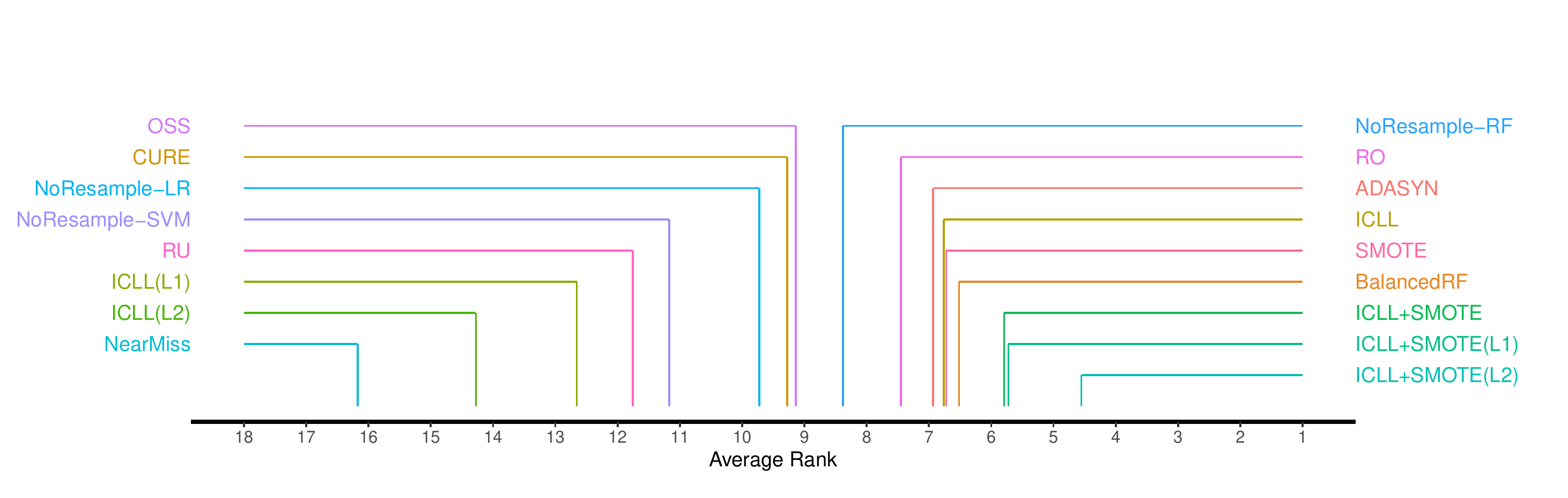}
    \caption{Average rank of each method using the difficult problems. Lower values denote better performance.}
    \label{fig:avg_rank_hm}
\end{figure}

% check d3flow abstract
\begin{figure}[h]
    \centering
    \includegraphics[width=\textwidth, trim=0cm 0cm 0cm 0cm, clip=TRUE]{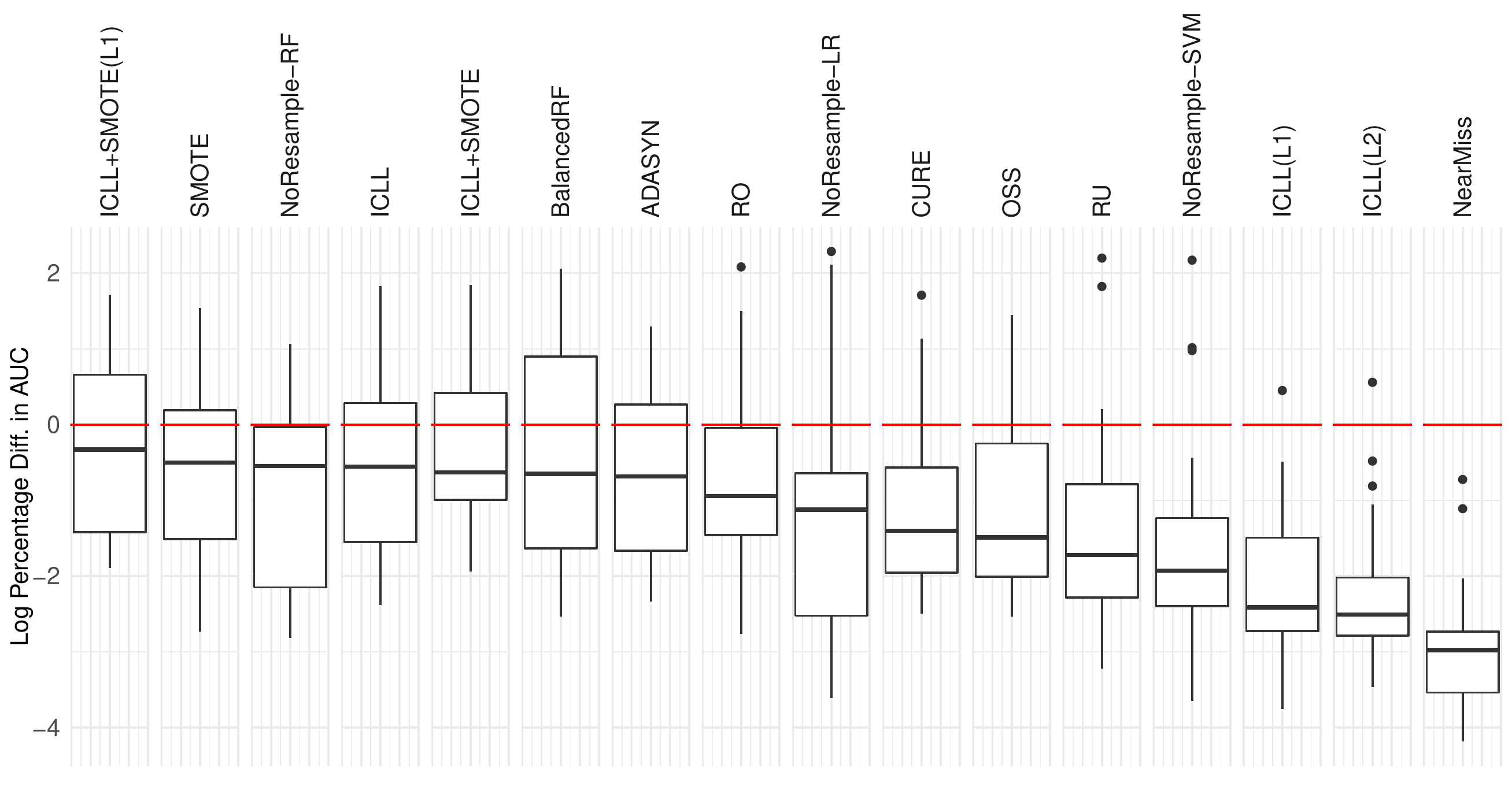}
    \caption{Boxplots showing the distribution of the log percentage difference between each method and \texttt{ICLL+SMOTE(L2)} using only the difficult problems}
    \label{fig:bp_pd_hm}
\end{figure}

Figure \ref{fig:avg_rank_hm} shows the average rank of each method across the difficult problems. The relative position of each method is similar. However, the variant of the proposed method without resampling (\texttt{ICLL}) shows a slightly worse average rank relative to \texttt{BalancedRF} and \texttt{SMOTE}. Notwithstanding, \texttt{ICLL+SMOTE(L2)} shows the best score overall.

\begin{figure}[h]
    \centering
    \includegraphics[width=\textwidth, trim=0cm 0cm 0cm 0cm, clip=TRUE]{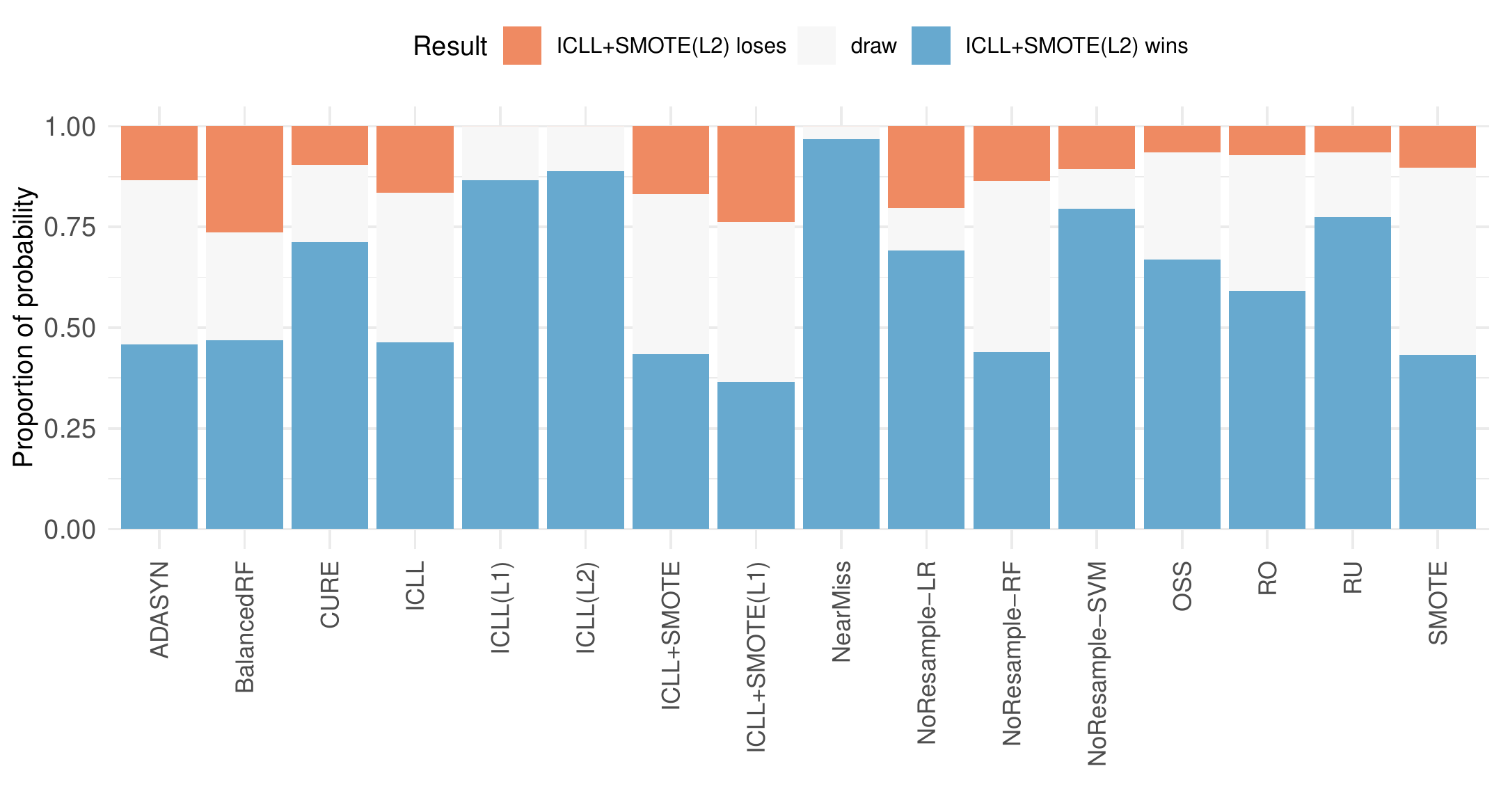}
    \caption{Paired comparisons between \texttt{ICLL+SMOTE(L2)} and each remaining approach. Each stacked barplot shows the probability of \texttt{ICLL+SMOTE(L2)} winning in blue (result below -1\%), drawing in light grey (result withing [-1\%, 1\%]), or losing in red (result above 1\%). This analysis considers only difficult problems.}
    \label{fig:prob_ratios_hm}
\end{figure}

\begin{figure}[h]
    \centering
    \includegraphics[width=\textwidth, trim=0cm 0cm 0cm 0cm, clip=TRUE]{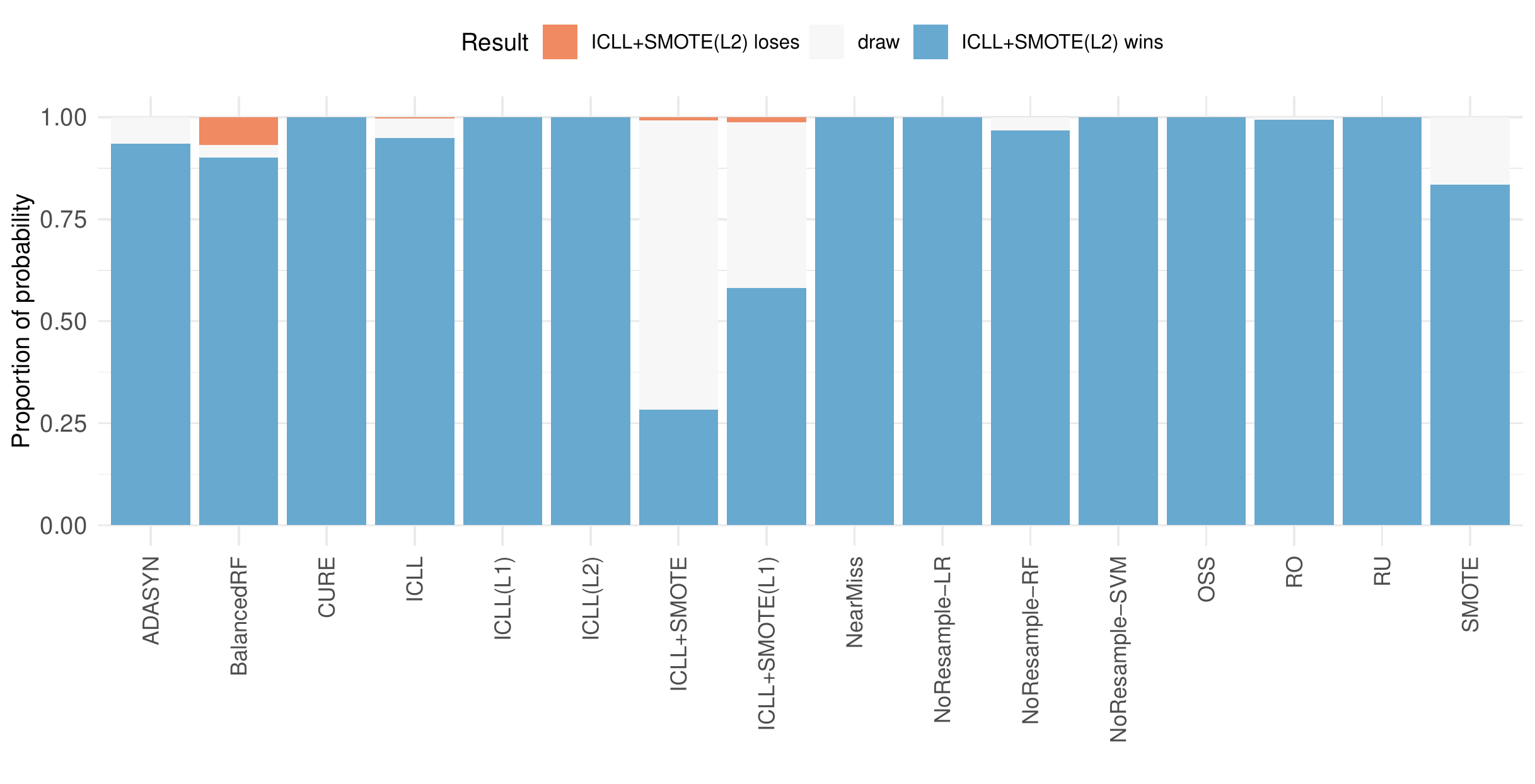}
    \caption{Paired comparisons between \texttt{ICLL+SMOTE(L2)} and each remaining approach using the Bayesian sign test with 1\% ROPE level. Each bar represents the proportion of probability of each outcome: the blue part denotes the probability that \texttt{ICLL+SMOTE(L2)} wins significantly; the red part represents the proportion of probability in which \texttt{ICLL+SMOTE(L2)} loses significantly; the grey area represents the probability that the difference in AUC is below 1\%.}
    \label{fig:bayes_hm}
\end{figure}

The distribution of the percentage difference in AUC is presented in Figure \ref{fig:bp_pd_hm}. Overall, for difficult problems the distribution of the percentage differences becomes even more favourable towards \texttt{ICLL+SMOTE(L2)}.
These conclusion can also be drawn from Figure \ref{fig:prob_ratios_hm}, which shows the probability of \texttt{ICLL+SMOTE(L2)} winning, losing, or drawing (percentage difference below 1\%). 
The results of the Bayesian signed-rank test are shown in Figure \ref{fig:bayes_hm}. In this subset of problems, \texttt{ICLL+SMOTE(L2)} shows statistically significant better performance when compared with all other methods, except for the variants \texttt{ICLL+SMOTE(L1)} and \texttt{ICLL+SMOTE}.
Indeed, when considering this subset of more difficult problems the advantage of the proposed method is enhanced and the probability of outperforming other methods increases considerably. 

\section{Discussion}\label{sec:discussion}

In the previous section we provided compelling evidence for the advantage of applying \texttt{ICLL} when tackling binary classification tasks with an imbalanced class distribution, especially when coupled with the \texttt{SMOTE} resampling method in the second layer.
Besides the gains in predictive performance, it is worth mentioning that the proposed method does not require any parameters besides the base learning algorithm, which in our case is a Random Forest. However, the best results were achieve when \texttt{ICLL} was coupled with \texttt{SMOTE}, which is not automated. Notwithstanding, in this work we applied \texttt{SMOTE} with its default configuration.
The hierarchical clustering procedure is automated. Specifically, the process of cutting the hierarchy and obtaining the cluster compositions is carried out using the heuristic described by Bellinger et al. \cite{bellinger2019cure}. 

One interesting thing we noted in the experiments is that \texttt{ICLL+SMOTE(L2)} shows a greater performance advantage in difficult data sets. These are problems in which the decision boundary is, in principle, more difficult to model. We believe that the concept of mixed group we introduced, and the subsequent stepwise approach based on layered learning, can be beneficial for these cases.

Layered learning approaches have been used for tackling classification problems. Notwithstanding, the layer definition is usually, to our knowledge, carried out manually -- either treating these as parameters to optimize or defined by domain experts. Therefore, automating the process of defining the layers within these approaches is a valuable contribution.

Our work is limited by the successful application of the hierarchical clustering procedure. To be more precise, the layers require the existence of both pure majority and mixed groups. Otherwise, these cannot be defined properly. Notwithstanding, during our experiments we found that majority groups were common because of the high prevalence of majority class instances. In 11 out of the 100 data sets, mixed groups could not be found. These represented easy data sets where the hierarchical clustering was able to perfectly split the two classes. Indeed, the classifier without any resampling method, or any other mechanism for dealing with class imbalance, achieved a perfect AUC score (1) in all those 11 data sets. In such cases, we do not need to go beyond the clustering analysis as this process indicated that an advanced classification strategy is not necessary. 

In terms of future developments, we outline two potential research directions: the first is a better exploitation of the hierarchy output by the clustering model. Specifically, it may be possible to devise different layer architectures depending on the result of the clustering analysis.
Second, we will attempt to apply \texttt{ICLL} to other predictive tasks, namely multi-class problems, one-class classification, or imbalanced regression tasks.

\section{Conclusions}\label{sec:conclusions}

We proposed a new approach for IBC problems, which is one of the most active research topics in machine learning.
The proposed approach models the data in a two-stage fashion according to a layered learning methodology \cite{stone2000layered}. The layers are automatically defined using hierarchical clustering analysis, and the class distribution of the resulting clusters.
We provided extensive empirical evidence which shows that the proposed approach leads to a better performance relatively to several state-of-the-art methods for IBC tasks. Contrary to the state of the art approaches to IBC, our proposal does not require tuning of any parameter - it is essentially parameter-less.

We believe that our method represents a promising direction towards modelling approaches which are not based on resampling the training data.

\section*{Declarations}

\begin{itemize}
    \item Funding: The work of L. Torgo was undertaken, in part, thanks to funding from the Canada Research Chairs program;
    
    \item Conflicts of interest/Competing interests: The authors have no relevant financial or non-financial interests to disclose;
    
    \item Ethics approval: Not applicable;
    
    \item Consent to participate: Not applicable;
    
    \item Consent for publication: Not applicable;
    
    \item Availability of data and material: All experiments and data are publicly available (c.f. footnote 1);
    
    \item Contributions: All authors contributed to writing and research.
\end{itemize}

\bibliographystyle{spmpsci}      % mathematics and physical sciences

\begin{thebibliography}{10}
\providecommand{\url}[1]{{#1}}
\providecommand{\urlprefix}{URL }
\expandafter\ifx\csname urlstyle\endcsname\relax
  \providecommand{\doi}[1]{DOI~\discretionary{}{}{}#1}\else
  \providecommand{\doi}{DOI~\discretionary{}{}{}\begingroup
  \urlstyle{rm}\Url}\fi

\bibitem{KEEL}
Keel data set repository.
\newblock \url{https://sci2s.ugr.es/keel/imbalanced.php#subA}.
\newblock Accessed: 2022-01-28

\bibitem{bellinger2019cure}
Bellinger, C., Branco, P., Torgo, L.: The cure for class imbalance.
\newblock In: International Conference on Discovery Science, pp. 3--17.
  Springer (2019)

\bibitem{benavoli2017time}
Benavoli, A., Corani, G., Dem{\v{s}}ar, J., Zaffalon, M.: Time for a change: a
  tutorial for comparing multiple classifiers through bayesian analysis.
\newblock The Journal of Machine Learning Research \textbf{18}(1), 2653--2688
  (2017)

\bibitem{branco2016survey}
Branco, P., Torgo, L., Ribeiro, R.P.: A survey of predictive modeling on
  imbalanced domains.
\newblock ACM Computing Surveys (CSUR) \textbf{49}(2), 1--50 (2016)

\bibitem{breiman2001random}
Breiman, L.: Random forests.
\newblock Machine learning \textbf{45}(1), 5--32 (2001)

\bibitem{cerqueira2020early}
Cerqueira, V., Torgo, L., Soares, C.: Early anomaly detection in time series: A
  hierarchical approach for predicting critical health episodes.
\newblock arXiv preprint arXiv:2010.11595  (2020)

\bibitem{chawla2002smote}
Chawla, N.V., Bowyer, K.W., Hall, L.O., Kegelmeyer, W.P.: Smote: synthetic
  minority over-sampling technique.
\newblock Journal of artificial intelligence research \textbf{16}, 321--357
  (2002)

\bibitem{chawla2003smoteboost}
Chawla, N.V., Lazarevic, A., Hall, L.O., Bowyer, K.W.: Smoteboost: Improving
  prediction of the minority class in boosting.
\newblock In: European conference on principles of data mining and knowledge
  discovery, pp. 107--119. Springer (2003)

\bibitem{chen2004using}
Chen, C., Liaw, A., Breiman, L., et~al.: Using random forest to learn
  imbalanced data.
\newblock University of California, Berkeley \textbf{110}(1-12), 24 (2004)

\bibitem{fernandez2018smote}
Fern{\'a}ndez, A., Garcia, S., Herrera, F., Chawla, N.V.: Smote for learning
  from imbalanced data: progress and challenges, marking the 15-year
  anniversary.
\newblock Journal of artificial intelligence research \textbf{61}, 863--905
  (2018)

\bibitem{fernandez2008study}
Fern{\'a}ndez, A., Garc{\'\i}a, S., del Jesus, M.J., Herrera, F.: A study of
  the behaviour of linguistic fuzzy rule based classification systems in the
  framework of imbalanced data-sets.
\newblock Fuzzy Sets and Systems \textbf{159}(18), 2378--2398 (2008)

\bibitem{fernandez2009hierarchical}
Fern{\'a}ndez, A., del Jesus, M.J., Herrera, F.: Hierarchical fuzzy rule based
  classification systems with genetic rule selection for imbalanced data-sets.
\newblock International Journal of Approximate Reasoning \textbf{50}(3),
  561--577 (2009)

\bibitem{galar2011review}
Galar, M., Fernandez, A., Barrenechea, E., Bustince, H., Herrera, F.: A review
  on ensembles for the class imbalance problem: bagging-, boosting-, and
  hybrid-based approaches.
\newblock IEEE Transactions on Systems, Man, and Cybernetics, Part C
  (Applications and Reviews) \textbf{42}(4), 463--484 (2011)

\bibitem{he2008adasyn}
He, H., Bai, Y., Garcia, E.A., Li, S.: Adasyn: Adaptive synthetic sampling
  approach for imbalanced learning.
\newblock In: 2008 IEEE international joint conference on neural networks (IEEE
  world congress on computational intelligence), pp. 1322--1328. IEEE (2008)

\bibitem{kubat1997addressing}
Kubat, M., Matwin, S., et~al.: Addressing the curse of imbalanced training
  sets: one-sided selection.
\newblock In: Icml, vol.~97, pp. 179--186. Citeseer (1997)

\bibitem{li2021autobalance}
Li, M., Zhang, X., Thrampoulidis, C., Chen, J., Oymak, S.: Autobalance:
  Optimized loss functions for imbalanced data.
\newblock Advances in Neural Information Processing Systems \textbf{34} (2021)

\bibitem{liu2008exploratory}
Liu, X.Y., Wu, J., Zhou, Z.H.: Exploratory undersampling for class-imbalance
  learning.
\newblock IEEE Transactions on Systems, Man, and Cybernetics, Part B
  (Cybernetics) \textbf{39}(2), 539--550 (2008)

\bibitem{mani2003knn}
Mani, I., Zhang, I.: knn approach to unbalanced data distributions: a case
  study involving information extraction.
\newblock In: Proceedings of workshop on learning from imbalanced datasets,
  vol. 126. ICML United States (2003)

\bibitem{moniz2021automated}
Moniz, N., Cerqueira, V.: Automated imbalanced classification via
  meta-learning.
\newblock Expert Systems with Applications \textbf{178}, 115,011 (2021)

\bibitem{murtagh2011methods}
Murtagh, F., Contreras, P.: Methods of hierarchical clustering.
\newblock arXiv preprint arXiv:1105.0121  (2011)

\bibitem{nickerson2001using}
Nickerson, A., Japkowicz, N., Milios, E.E.: Using unsupervised learning to
  guide resampling in imbalanced data sets.
\newblock In: International Workshop on Artificial Intelligence and Statistics,
  pp. 224--228. PMLR (2001)

\bibitem{pedregosa2011scikit}
Pedregosa, F., Varoquaux, G., Gramfort, A., Michel, V., Thirion, B., Grisel,
  O., Blondel, M., Prettenhofer, P., Weiss, R., Dubourg, V., et~al.:
  Scikit-learn: Machine learning in python.
\newblock the Journal of machine Learning research \textbf{12}, 2825--2830
  (2011)

\bibitem{ribeiro2021layered}
Ribeiro, B., Cerqueira, V., Santos, R., Gamboa, H.: Layered learning for acute
  hypotensive episode prediction in the icu: An alternative approach.
\newblock In: 2021 International Conference on e-Health and Bioengineering
  (EHB), pp. 1--4. IEEE (2021)

\bibitem{sharma2012anomaly}
Sharma, S., Bellinger, C., Japkowicz, N., Berg, R., Ungar, K.: Anomaly
  detection in gamma ray spectra: A machine learning perspective.
\newblock In: 2012 IEEE Symposium on Computational Intelligence for Security
  and Defence Applications, pp. 1--8. IEEE (2012)

\bibitem{stone2000layered}
Stone, P., Veloso, M.: Layered learning.
\newblock In: European Conference on Machine Learning, pp. 369--381. Springer
  (2000)

\bibitem{sutton1999between}
Sutton, R.S., Precup, D., Singh, S.: Between mdps and semi-mdps: A framework
  for temporal abstraction in reinforcement learning.
\newblock Artificial intelligence \textbf{112}(1-2), 181--211 (1999)

\bibitem{tomek1976two}
Tomek, I., et~al.: Two modifications of cnn.  (1976)

\bibitem{wang2009diversity}
Wang, S., Yao, X.: Diversity analysis on imbalanced data sets by using ensemble
  models.
\newblock In: 2009 IEEE symposium on computational intelligence and data
  mining, pp. 324--331. IEEE (2009)

\bibitem{wu2007local}
Wu, J., Xiong, H., Wu, P., Chen, J.: Local decomposition for rare class
  analysis.
\newblock In: Proceedings of the 13th ACM SIGKDD international conference on
  Knowledge discovery and data mining, pp. 814--823 (2007)

\end{thebibliography}

\end{document}